\documentclass[a4paper]{article}

\usepackage[english]{babel}
\usepackage[latin1]{inputenc}
\usepackage[T1]{fontenc}

\usepackage{yfonts}

\usepackage{mathtools}
\usepackage[pdftex]{graphicx}
\usepackage{epstopdf}
\usepackage{makeidx}
\usepackage{nicefrac}
\usepackage{multicol}
\usepackage[bottom]{footmisc}
\usepackage{comment}
\usepackage[margin=15pt,font=small,labelfont=bf,labelsep=endash,aboveskip=-3pt,belowskip=-8pt]{caption}
\usepackage[most]{tcolorbox}
\usepackage{forest}
\usepackage[e]{esvect}
\usepackage{pdflscape}
\usepackage{pdfpages}
\usepackage{comment}
\usepackage{amsfonts}
\usepackage{amssymb}
\usepackage{amsmath}
\usepackage{amsthm}
\usepackage{algorithm2e}

\allowdisplaybreaks

\usepackage{url}
\usepackage{mlmodern}
\usepackage[T1]{fontenc}

\usepackage[small,euler-hat-accent]{eulervm}

\DeclareMathSizes{10}{9.5}{7}{7}

\usepackage{epsfig}
\usepackage{color}

\newcommand {\real}{\mathbb{R}}

\newcommand {\complexo}{\mathbb{C}}
\newcommand {\nat}{\mathbb{N}}

\usepackage{bm}
\newcommand{\negmath}[1]{\mathbold{{#1}}}

\newtcolorbox[auto counter, number within=chapter]{example}[1][]{%
	enhanced jigsaw, 
	breakable=true,
	before skip=7pt,
	after skip=7pt,
	left=0mm,
	right=0mm,
	top=1mm,
	bottom=1.5mm,
	boxsep=1.5mm,
	borderline west={1pt}{0pt}{black!23!white}, 
	sharp corners, 
	boxrule=0pt, 
	fonttitle={\small\bfseries},
	coltitle={black},  
	opacityfill=0,
	fontupper=\small,
	title={Example~\thetcbcounter.\ },  
	attach title to upper, 
	#1
}

\newcounter{th}
\newtcbtheorem[list inside=mteo]{mteo}{Theorem}
{
	enhanced jigsaw,
	arc=0mm,outer arc=.5mm,
	breakable=true,
	colback=black!1!white,
	before skip=5pt,
	after skip=5pt,
	leftrule=0pt,
	rightrule=0pt,
	coltitle=black,
	top=1mm,
	bottom=1.5mm,
	left=0mm,
	right=0mm,
	colframe=black!23!white,
	fonttitle=\bfseries,
	separator sign dash,
	theorem style=standard,
	fontupper=\itshape,
	titlerule=0mm,
	separator sign dash,
	before title={\stepcounter{th}}
}{teo}

\newtcbtheorem[number within=th,list inside=mteo]{mcoro}{Corollary}
{
	enhanced jigsaw,
	arc=0mm,outer arc=.5mm,
	breakable=true,
	colback=black!1!white,
	before skip=5pt,
	after skip=5pt,
	leftrule=0pt,
	rightrule=0pt,
	coltitle=black,
	top=0.5mm,
	bottom=0.5mm,
	left=0mm,
	right=0mm,
	colframe=black!23!white,
	fonttitle=\bfseries,
	separator sign dash,
	theorem style=standard,
	fontupper=\itshape,
	titlerule=0mm,
	separator sign dash
}{cor}

\tcolorboxenvironment{proof}{	
	enhanced jigsaw, 
	breakable=true,
	before skip=5pt,
	after skip=5pt,
	left=1mm,
	right=0mm,
	top=0.5mm,
	bottom=0.5mm,
	boxsep=1mm,
	borderline west={1.5pt}{0pt}{black!23!white}, 
	sharp corners, 
	boxrule=0pt, 
	fonttitle={\small\bfseries},
	coltitle={black},  
	opacityfill=0,
	attach title to upper, 
}

\newcommand {\map}[3]{#1 : #2 \mapsto #3}

\newcommand {\dete}[1]{\mathrm{det}\,(#1)}

\newcommand {\vto}[1]{\negmath{#1}}

\newcommand {\vun}[1]{\negmath{\hat{#1}}}

\newcommand {\vtf}[1]{\pmb{#1}}

\newcommand {\mat}[1]{\mathit{#1}}

\newcommand {\tnr}[1]{\vtf{#1}}

\title{Tensor-Based Foundations of Ordinary Least Squares and Neural Network Regression Models}

\author{Roberto Dias Algarte\thanks{robertodias70@outlook.com}}

\begin{document}

\RestyleAlgo{ruled}

\maketitle

\begin{abstract}
\noindent This article introduces a novel approach to the mathematical development of Ordinary Least Squares and Neural Network regression models, diverging from traditional methods in current Machine Learning literature. By leveraging Tensor Analysis and fundamental matrix computations, the theoretical foundations of both models are meticulously detailed and extended to their complete algorithmic forms. The study culminates in the presentation of three algorithms, including a streamlined version of the Backpropagation Algorithm for Neural Networks, illustrating the benefits of this new mathematical approach.  
\end{abstract}

\section{Vectors, Tensors and Matrices}
The following sections of this article require some important mathematical concepts and notations that need to be addressed here in this section. However, we assume the reader to be proficient on the basic and intermediate topics of Linear Algebra and Calculus, which will not be covered. 

If $U$ is a set and $f$ a rule described by an expression like $f(x)=\alpha x+\beta$, given arbitrary elements $a,b\in U$, the pair $(U,f)$ is called a function if and only if whenever $a=b$ then $f(a)=f(b)$. In this context, $U$ is said to be the domain of the function and $\map{f}{U}{V}$ the mapping of $f$ from $U$ to codomain $V$, from which the set $Im_f\subseteq V$, constituted by all $f(a)$, is called the image of $f$. The element $f(a)\in Im_f$ is commonly said to be the value of $f$ at $a$.

Now, let $\map{f}{U}{V}$ be a generic mapping where $f$ is unknown and the elements in the sets $\widetilde{U}:=\{u^1,\cdots,u^p\}\subseteq U$ and $\widetilde{V}:=\{f(u^1),\cdots,f(u^p)\}\subseteq V$, $p\in\nat^*$, are known. In simple terms, the solution of a \emph{supervised machine learning} problem is an approximation function $\widetilde{f}\approx f$ obtained from the set of pairs $\mathcal{T}=\{ (u^1,f(u^1),\cdots, (u^p,f(u^p)\}$, called the \emph{training set}. This problem is said to be a \emph{classification problem} when the codomain $V$ is a discrete set, and a \emph{regression problem} otherwise.  The elements of $U$ are usually called \emph{features} and its values in $V$ are said to be \emph{labels} or \emph{targets}. An \emph{unsupervised machine learning problem}, not covered in this article, seeks to find $\widetilde{f}\approx f$ from $\mathcal{T}=\widetilde{U}$; in other words, in this problem, no targets are present.  

The elements of the real $\real$ or the complex $\complexo$ sets are called scalars, being magnitudes of order zero. If the sets under study are vector or tensor spaces, their elements are respectively magnitudes of order one and grater than one. Here, we shall work only with real Hilbert spaces, which are complete normed inner product vector spaces defined by real scalars where the inner product induces the norm. Since every $n$-dimensional Hilbert space $\tnr{\mathcal{U}}^{n}$ has orthonormal basis $\{\vun{u}_1,\cdots,\vun{u}_n\}$, an arbitrary vector $\vto{u}\in \tnr{\mathcal{U}}^{n}$ can be described by 
\begin{equation}
\vto{u}=\sum_{i=1}^n u_i\vun{u}_i\,,
\end{equation}
where the scalars $u_i$ are the coordinates of $\vto{u}$ on the basis under consideration. For our practical purposes, given arbitrary vectors $\vto{u}\in\tnr{\mathcal{U}}^{n}$ and $\vto{v}\in\tnr{\mathcal{V}}^{m}$, an element $\vto{u}\otimes\vto{v}$ of a tensor space $\tnr{\mathcal{U}}^{n}\otimes\tnr{\mathcal{V}}^{m}$ is called a tensor and can be decomposed, similarly to a vector, by 
\begin{equation}\label{eq:compsTensor}
\vto{u}\otimes\vto{v}=\sum_{i=1}^n\sum_{i=1}^m (\vto{u}\otimes\vto{v})_{ij}\vun{u}_i\otimes\vun{v}_i\,,
\end{equation}
where the scalars $(\vto{u}\otimes\vto{v})_{ij}$ are the coordinates of $\vto{u}\otimes\vto{v}$ on the basis of tensors	
\begin{equation}
\{\vun{u}_1\otimes\vun{v}_1,\vun{u}_1\otimes\vun{v}_2,\cdots,\vun{u}_1\otimes\vun{v}_m,\vun{u}_2\otimes\vun{v}_1,\cdots,\vun{u}_n\otimes\vun{v}_m\}\,.
\end{equation}
Thereby, $\tnr{\mathcal{U}}^{n}\otimes\tnr{\mathcal{V}}^{m}$ is a tensor space of dimension $nm\,$. Again, for the purposes of this study, the practical goal of symbol $\otimes$, called tensor product, is here to compose two magnitudes of order one (vectors) to generate a magnitude of order two (tensor). Therefore, a tensor $\vto{x}_1\otimes\cdots\otimes\vto{x}_p$ has order $p\geqslant 1$. A tensor of order $2p$ is called identity tensor if 
\begin{equation}\label{eq:identity}
\tnr{I}^{2p} = \sum_{i_1=1}^{n_1} \cdots\sum_{i_p=1}^{n_p} \vun{u}_{i_1}^{(1)}\otimes\cdots\otimes\vun{u}_{i_p}^{(p)}\otimes\vun{u}_{i_1}^{(1)}\otimes\cdots\otimes\vun{u}_{i_p}^{(p)}\,,
\end{equation}
where the set $\{\vun{u}_{1}^{(j)},\cdots,\vun{u}_{n_j}^{(j)}\}$ is an orthonormal basis in each order $j$.

It is quite convenient to represent vector $\vto{u}$ and second order tensor $\vto{u}\otimes\vto{v}$ by matrices the following way:
\begin{equation}
[\vto{u}]:=\begin{bmatrix}
u_1 & \cdots & u_n\\
\end{bmatrix}^T\,;
\end{equation}
\begin{equation}
[\vto{u}\otimes\vto{v}]:=
\begin{bmatrix}
(\vto{u}\otimes\vto{v})_{11} & \cdots & (\vto{u}\otimes\vto{v})_{1m}\\
\vdots & \ddots & \vdots\\
(\vto{u}\otimes\vto{v})_{n1} & \cdots & (\vto{u}\otimes\vto{v})_{nm}
\end{bmatrix}\,.
\end{equation}
For the purposes of this article, it is also useful to consider the so called \emph{Frobenius Inner Product} of matrices, namely, 
\begin{equation}
[\vto{u}_1\otimes\vto{v}_1]:[\vto{u}_2\otimes\vto{v}_2]:=\sum_{i=1}^m\sum_{i=1}^n[\vto{u}_1\otimes\vto{v}_1]_{ij}[\vto{u}_2\otimes\vto{v}_2]_{ij}\,,
\end{equation}
where $\vto{u}_1,\vto{u}_2\in\tnr{\mathcal{V}}^{m}$ and $\vto{v}_1,\vto{v}_2\in\tnr{\mathcal{U}}^{n}$, from which the \emph{Frobenius Norm} is defined:
\begin{equation}
\|[\vto{u}\otimes\vto{v}]\|= \sqrt{[\vto{u}\otimes\vto{v}]:[\vto{u}\otimes\vto{v}]}\,.
\end{equation}
The order-one inner product of $\vto{u}_1\otimes\vto{v}_1$ with a vector $\vto{v}_2\in \tnr{\mathcal{V}}^{m}$ is the vector 
\begin{equation}\label{eq:prodIntTv}
(\vto{u}_1\otimes\vto{v}_1)\odot\vto{v}_2=(\vto{v}_1\cdot\vto{v}_2)\vto{u}_1\,,
\end{equation}
and with a tensor $(\vto{v}_2\otimes\vto{u}_2)\in\tnr{\mathcal{V}}^{m}\otimes\tnr{\mathcal{U}}^{n}$ is 
\begin{equation}
(\vto{u}_1\otimes\vto{v}_1)\odot(\vto{v}_2\otimes\vto{u}_2)=(\vto{v}_1\cdot\vto{v}_2)(\vto{u}_1\otimes\vto{u}_2)\,,
\end{equation}
which is a tensor that belongs to $\tnr{\mathcal{U}}^{n}\otimes\tnr{\mathcal{U}}^{n}$. Note that the usual inner product, represented as a simple dot symbol, is actually the order-one inner product of two order-one magnitudes and results a scalar. The order-two inner product of two order-two magnitudes also results a scalar by definition
\begin{equation}
(\vto{u}_1\otimes\vto{v}_1)\odot_2(\vto{u}_2\otimes\vto{v}_2)=(\vto{u}_1\cdot\vto{u}_2)(\vto{v}_1\cdot\vto{v}_2)\,.
\end{equation}
From the definitions above, the following properties are straightforwardly obtained:
\begin{itemize}
  \item[i.] $\alpha(\vto{u}\otimes\vto{v})=(\alpha\vto{u})\otimes\vto{v}=\vto{u}\otimes(\alpha\vto{v})$ for all $\alpha\in\real$;   
  \item [ii.] $\vto{u}\otimes(\vto{v}_1+\vto{v}_2)=\vto{u}\otimes\vto{v}_1+\vto{u}\otimes\vto{v}_2$;
  \item [iii.] $(\vto{u}_1+\vto{u}_2)\otimes\vto{v}=\vto{u}_1\otimes\vto{v}+\vto{u}_2\otimes\vto{v}$;
  \item [iv.] $(\vto{u}_1\otimes\vto{u}_2)_{ij}=(\vto{u}_1)_i(\vto{u}_2)_i$;
  \item [v.] $[(\vto{u}\otimes\vto{v}_1)\odot\vto{v}_2]=[\vto{u}\otimes\vto{v}_1][\vto{v}_2]$;
  \item [vi.] $[(\vto{u}_1\otimes\vto{v}_1)\odot(\vto{v}_2\otimes\vto{u}_2)]=[\vto{u}_1\otimes\vto{v}_1][\vto{v}_2\otimes\vto{u}_2]$;
  \item [vii.] $(\vto{u}_1\otimes\vto{v}_1)\odot_2(\vto{u}_2\otimes\vto{v}_2)=[\vto{u}_1\otimes\vto{v}_1]:[\vto{u}_2\otimes\vto{v}_2]$;
    \item [viii.] $\tnr{I}^2\odot\vto{u}=\vto{u}\odot\tnr{I}^2=\vto{u}$.
\end{itemize}
When it is convenient, we shall talk of a tensor $\tnr{T}\in\tnr{\mathcal{U}}^{n}\otimes\tnr{\mathcal{V}}^{m}$ instead of writing the tensor product $\vto{u}\otimes\vto{v}$ of two vectors.

A tensor $(\vto{u}\otimes\vto{v})^T$ is called the transpose of tensor $(\vto{u}\otimes\vto{v})$ if   
\begin{equation}
(\vto{u}\otimes\vto{v})^T=\vto{v}\otimes\vto{u}\,,
\end{equation}
and $(\vto{u}\otimes\vto{v})^{-1}$ its inverse if $n=m$ and   
\begin{equation}
(\vto{u}\otimes\vto{v})\odot(\vto{u}\otimes\vto{v})^{-1}=\tnr{I}^2\,,
\end{equation}

For a mapping  $\map{\vtf{f}}{\tnr{U}}{\tnr{V}}$, where $\tnr{U}$ and $\tnr{V}$ are subsets of $\tnr{\mathcal{U}}^{n}$, $\vtf{f}$ is called a vector function. Sometimes, it is useful to describe a vector function in terms of its \emph{coordinate functions}, namely,
\begin{equation}
	\vtf{f}(\vto{x}) = \sum_{i=1}^{n} f_i(\vto{x})\vun{v}_i\,,
\end{equation}
where $f_i$ is a scalar-valued vector function. In the case of 
\begin{equation}
\vtf{f}(\alpha\vto{u}_1+\beta\vto{u}_2) = \alpha\vtf{f}(\vto{u}_1) + \beta\vtf{f}(\vto{u}_2)\,,\,\forall\, \vto{u}_1,\vto{u}_2\in\tnr{U}\,,\,\alpha,\beta\in\real\,, 
\end{equation}
$\vtf{f}$ is called linear. An important property of linear functions is that $\vtf{f}(\vto{0})=\vto{0}$.

\section{Ordinary Least Squares}
Considering the set $\tnr{U}$ a $n$-dimensional subspace of a real Hilbert space $\tnr{\mathcal{U}}^{n+1}$ and the set $\tnr{V}$ a subspace of $\tnr{\mathcal{V}}^{m}$, where $n,m\in\nat^*$, let 
\begin{equation}
\pmb{\mathcal{T}}=\{ (\vto{u}^1,\vtf{f}(\vto{u}^1)),\cdots, (\vto{u}^p,\vtf{f}(\vto{u}^p))\}\,,\, p\in\nat^*,
\end{equation}
be the training set of a regression problem where $\map{\vtf{f}}{\tnr{U}}{\tnr{V}}$. If $\{\vun{u}_1,\cdots,\vun{u}_{n+1}\}$ and $\{\vun{v}_1,\cdots,\vun{v}_{m}\}$ are orthonormal bases of $\tnr{\mathcal{U}}^{n+1}$ and $\tnr{\mathcal{V}}^{m}$ respectively, we can decompose the training features and targets through  
\begin{alignat}{3} 
\vto{u}^j=\sum_{i=1}^{n}u^j_i\vun{u}_i&\qquad \text{and}\qquad & \vtf{f}(\vto{u}^j)=\sum_{i=1}^{m}f_{i}(\vto{u}^j)\vun{v}_i\,,
\end{alignat} 
where $j=1,\cdots,p$. In this context, since an element of $\tnr{U}$ cannot be a scalar, it is commonly called a \emph{feature-vector}, and its value a \emph{target-vector}. Considering arbitrary orthonormal bases of $\tnr{\mathcal{U}}^{n+1}$ and $\tnr{\mathcal{V}}^{m}$, the goal of the linear approach for a regression problem is to find a tensor $\tnr{B}\in \tnr{\mathcal{V}}^{m}\otimes\tnr{\mathcal{U}}^{n+1}$ such that 
\begin{equation}
\widetilde{\vtf{f}}(\vto{x})= (\sum_{i=1}^{m}\sum_{j=1}^{n+1} B_{ij}\vun{v}_i\otimes\vun{u}_j)\odot(\vun{u}_{n+1}+ \sum_{j=1}^{n} x_{j}\vun{u}_j)\,.
\end{equation}
Now, we use a little bit of mathematical sagacity to obtain a linear relation from the previous rule. By assuming the existence of a subset $\overline{\tnr{U}}\subset\tnr{\mathcal{U}}^{n+1}$ constituted by the vectors $\vun{u}_{n+1}+\vto{u}$, where $\vto{u}\in\tnr{U}$, a mapping $\map{\widetilde{\vtf{g}}}{\overline{\tnr{U}}}{\tnr{V}}$ can be defined in such a way that $\widetilde{\vtf{g}}(\vun{u}_{n+1}+\vto{u})=\widetilde{\vtf{f}}(\vto{u})=\tnr{B}\odot(\vun{u}_{n+1}+\vto{u})$ for all $\vto{u}\in\tnr{U}$. Therefore, $\widetilde{\vtf{g}}$ results a linear function with rule 
\begin{equation}
\widetilde{\vtf{g}}(\vto{x})= \tnr{B}\odot\vto{x}\,.
\end{equation}
In this context, considering an arbitrary vector $\vto{u}\in\tnr{U}$, we specify $\overline{\vto{u}}\in\overline{\tnr{U}}$ such that $\overline{\vto{u}}:=\vun{u}_{n+1}+\vto{u}$ and then for each dimension $i$ of $\tnr{\mathcal{V}}^{m}$, 	
\begin{equation}
\widetilde{g}_i(\overline{\vto{u}})= B_{i1}\overline{u}_1+\cdots+B_{in}\overline{u}_n+B_{i(n+1)}\,.
\end{equation}
Considering the previous conditions, from the adapted training set
\begin{equation}
\pmb{\mathcal{T}}=\{ (\overline{\vto{u}}^1,{\vtf{f}}({\vto{u}}^1)),\cdots, (\overline{\vto{u}}^p,{\vtf{f}}({\vto{u}}^p))\}\,,
\end{equation}
the \emph{Ordinary Least Squares Method} (OLS) defines the scalar valued tensor mapping $\map{\tnr{\psi}}{\tnr{\mathcal{V}}^{m}\otimes\tnr{\mathcal{U}}^{n+1}}{\real^+}$ with function rule 
\begin{equation}\label{eq:MMQ}
\tnr{\psi}(\tnr{X})=\sum_{k=1}^{p}\|\underbrace{\vtf{f}(\vto{u}^k)-\tnr{X}\odot\overline{\vto{u}}^k}_{\vtf{r}_k(\tnr{X})}\|^2\,,
\end{equation}
where $\|\bullet\|$ is the Euclidean norm and $\vtf{r}_k$ is a vector valued tensor function called the \emph{residual} between the $k$-th training target-vector and its approximate value in the linear regression model. In other words,
\begin{equation}\label{eq:MMQR}
\tnr{\psi}(\tnr{X})=\sum_{k=1}^{p}\vtf{r}_k(\tnr{X})\cdot{\vtf{r}_k(\tnr{X})}\,.
\end{equation}
The OLS then proposes to obtain a tensor $\tnr{B}\in\tnr{\mathcal{V}}^{m}\otimes\tnr{\mathcal{U}}^{n+1}$ such that the error 
\begin{equation}
\tnr{\psi}(\tnr{B})=\min_{\tnr{X}\in\tnr{\mathcal{V}}^{m}\otimes\tnr{\mathcal{U}}^{n+1}}\tnr{\psi}(\tnr{X})\,.
\end{equation}
Since rule \eqref{eq:MMQ} is the sum of quadratic terms, it follows that tensor 
\begin{equation}\label{eq:nabla}
\nabla\tnr{\psi}(\tnr{B})=\tnr{0}\,,
\end{equation}
which is called the gradient of $\tnr{\psi}$ at $\tnr{B}$. In our approach, given a non zero arbitrary tensor $\tnr{H}\in\tnr{\mathcal{V}}^{m}\otimes\tnr{\mathcal{U}}^{n+1}$, the derivative of $\tnr{\psi}$ on direction $\tnr{H}$ is 
\begin{equation}\label{eq:defGrad}
[\tnr{\psi}'(\tnr{X})](\tnr{H})=\lim_{\alpha\to 0} \dfrac{\tnr{\psi}(\tnr{X}+\alpha\tnr{H})-\tnr{\psi}(\tnr{X})}{\alpha}\,,  
\end{equation}
from which the Tensor Calculus classically defines that
\begin{equation}
[\tnr{\psi}'(\tnr{X})](\tnr{H})=\tnr{H}\odot_2\nabla\tnr{\psi}(\tnr{X})\,.  
\end{equation}
From this definition, condition \eqref{eq:nabla} and rule \eqref{eq:MMQ}, we can develop the following:
\begin{align}
0=&[\tnr{\psi}'(\tnr{B})](\tnr{H})\nonumber\\  
=&\lim_{\alpha\to0}1/\alpha[\sum_{k=1}^p\vtf{r}_k(\tnr{B}+\alpha\tnr{H})\cdot\vtf{r}_k(\tnr{B}+\alpha\tnr{H})-\vtf{r}_k(\tnr{B})\cdot\vtf{r}_k(\tnr{B})]\nonumber\\
=&\lim_{\alpha\to0}1/\alpha[\sum_{k=1}^p(\vtf{r}_k(\tnr{B})-\alpha\tnr{H}\odot\overline{\vto{u}}^k)\cdot(\vtf{r}_k(\tnr{B})-\alpha\tnr{H}\odot\overline{\vto{u}}^k)-\vtf{r}_k(\tnr{B})\cdot\vtf{r}_k(\tnr{B})]\nonumber\\
=&\lim_{\alpha\to0}1/\alpha[\sum_{k=1}^p-2\alpha\vtf{r}_k(\tnr{B})\cdot\tnr{H}\odot\overline{\vto{u}}^k + \alpha^2\tnr{H}\odot\overline{\vto{u}}^k\cdot\tnr{H}\odot\overline{\vto{u}}^k]\nonumber\\
=&\sum_{k=1}^p-2\vtf{r}_k(\tnr{B})\cdot\tnr{H}\odot\overline{\vto{u}}^k\nonumber\\
=&-2\tnr{H}\odot_2\sum_{k=1}^p\vtf{r}_k(\tnr{B})\otimes\overline{\vto{u}}^k\nonumber\\
=&\sum_{k=1}^p\vtf{r}_k(\tnr{B})\otimes\overline{\vto{u}}^k\label{eq:develop}\,,
\end{align}
since direction $\tnr{H}$ is not zero by definition. Developing the residual in the last equality, we arrive at what is usually called the \emph{normal equation}, namely  
\begin{equation}\label{eq:tensB}
\tnr{B}\odot\sum_{k=1}^p \overline{\vto{u}}^k\otimes\overline{\vto{u}}^k=\sum_{k=1}^p\vtf{f}(\vto{u}^k)\otimes\overline{\vto{u}}^k\,, 
\end{equation}
which will be arranged in matrix form in order to easily calculate the unknown tensor $\tnr{B}$ analytically. Given the matrices
\begin{alignat}{3}
\mat{X}:=\begin{bmatrix}
{u}^1_1 & \cdots & {u}^1_n & 1\\
\vdots & \ddots & \vdots & \vdots\\
{u}^p_1 & \cdots & {u}^p_{n} & 1
\end{bmatrix}&\qquad \text{and}\qquad & \mat{Y}:=\begin{bmatrix}
{f}_1({\vto{u}}^1) & \cdots & {f}_1({\vto{u}}^p)\\
\vdots & \ddots & \vdots\\
{f}_m({\vto{u}}^1) & \cdots & {f}_m({\vto{u}}^p)
\end{bmatrix}\,,
\end{alignat}
by considering the properties of tensors presented in the previous section, it is straightforward to obtain that
\begin{alignat}{3} 
[\sum_{k=1}^p \overline{\vto{u}}^k\otimes\overline{\vto{u}}^k]=\mat{X}^T\mat{X}&\qquad \text{and}\qquad & [\sum_{k=1}^p\vtf{f}(\vto{u}^k)\otimes\overline{\vto{u}}^k] = \mat{Y}\mat{X}\,.
\end{alignat}     
Therefore, from the normal equation \eqref{eq:tensB}, if the square matrix $\mat{Z}:=\mat{X}^T\mat{X}$ is invertible, that is, if $\dete{\mat{Z}}\neq 0$, we can write that the $m\times (n+1)$ matrix
\begin{equation}\label{eq:analytic}
[\tnr{B}]=\mat{Y}\mat{X}\mat{Z}^{-1}\,. 
\end{equation}
An algorithmic scheme for solving this equation is rather simple, as shown below in Algorithm 1.  
\begin{algorithm}
\caption{Analytical calculation of tensor $\tnr{B}$}
\KwData{Training set $\pmb{\mathcal{T}}$}
\KwResult{$[\tnr{B}]$}
$\mat{X} \gets \begin{bmatrix}
u^k_1 & \cdots & u^k_n & 1\\
\end{bmatrix}$ stacked vertically\;
$\mat{Z} \gets \mat{X}^T\mat{X}$\;
  \If{$\dete{\mat{Z}}\neq 0$}{
    $\mat{Y} \gets \begin{bmatrix}
      {f}_1({\vto{u}}^k) & \cdots & {f}_m({\vto{u}}^k)\\
      \end{bmatrix}^T$ stacked horizontally\;
    $[\tnr{B}] \gets \mat{Y}\mat{X}\mat{Z}^{-1}$\;
  }
\end{algorithm}

\paragraph{The Gradient Descent Method.} The main limitation of the previous analytical equation for tensor $\tnr{B}$ is that $\mat{Z}$ must be non singular, a limitation that may become a numerical problem if this matrix is nearly singular, be this condition caused by roundoff errors or not. Therefore, the stability of the previous analytical solution depends on the level of singularity of the matrix $\mat{Z}$. An effective iterative strategy that avoids this restriction in finding the coefficient $\tnr{B}$ of a linear regression problem is called the \emph{Gradient Descent Method} (GD), which we present as follows. Let $\map{\tnr{\varphi}}{\tnr{\mathcal{U}}^{q}\otimes\tnr{\mathcal{V}}^{s}}{\real}$, where $q,s\in\nat^*$, be a differentiable mapping and $\tnr{T}_0\in\tnr{\mathcal{U}}^{q}\otimes\tnr{\mathcal{V}}^{s}$ an arbitrary tensor from which a local nearby minimum $\tnr{\varphi}(\tnr{T})$, $\tnr{T}\in\tnr{\mathcal{U}}^{q}\otimes\tnr{\mathcal{V}}^{s}$, is considered to exist. In this context, the GD states that a sequence $\tnr{T}_0,\tnr{T}_1,\cdots$ converges to $\tnr{T}$ if, at every iteration $t\in 0,1,...$,  
\begin{equation}\label{eq:gdm}
\tnr{T}_{t+1} = \tnr{T}_{t} - \gamma_t\nabla\tnr{\varphi}(\tnr{T}_t)\,,
\end{equation}
where $\gamma_t\in\real^+$ is a scalar called the \emph{learning rate}. In other words, if the difference $(\tnr{T}_{t+1} - \tnr{T}_{t})$ is a convenient multiple of the descent gradient $-\nabla\tnr{\varphi}(\tnr{T}_t)$, then the scalar sequence $\tnr{\varphi}(\tnr{T}_0),\tnr{\varphi}(\tnr{T}_1),\cdots$ converges to a local minimum. This convenient multiplicity, the learning rate, can be specified to be a constant scalar, but here we shall use the so called \emph{Barzilai-Borwein method}, which states that the convergence of the GD is faster if, for every iteration $t>0$, 
\begin{equation}\label{eq:bbm}
\gamma_t = \dfrac{|(\tnr{T}_{t}-\tnr{T}_{t-1})\odot_2(\nabla\tnr{\varphi}(\tnr{T}_t)-\nabla\tnr{\varphi}(\tnr{T}_{t-1}))|}{\|\nabla\tnr{\varphi}(\tnr{T}_t)-\nabla\tnr{\varphi}(\tnr{T}_{t-1})\|^2}\,.
\end{equation}

Since the function $\tnr{\psi}$, defined by the rule \eqref{eq:MMQR}, observes the conditions imposed on $\tnr{\varphi}$, let's apply the GD to the OLS problem. From development \eqref{eq:develop}, definition \eqref{eq:defGrad} and their conditions, for any direction $\tnr{H}$, we can write that
\begin{equation*}
\tnr{H}\odot_2\nabla\tnr{\psi}(\tnr{B})= -2\tnr{H}\odot_2\sum_{k=1}^p\vtf{r}_k(\tnr{B})\otimes\overline{\vto{u}}^k\,,
\end{equation*}
from which equality 
\begin{equation*}
\nabla\tnr{\psi}(\tnr{B})= 2(\tnr{B}\odot\sum_{k=1}^p \overline{\vto{u}}^k\otimes\overline{\vto{u}}^k-\sum_{k=1}^p\vtf{f}(\vto{u}^k)\otimes\overline{\vto{u}}^k)
\end{equation*}
is obtained by considering the normal equation \eqref{eq:tensB}. From this equality, recalling the definitions of matrices $\mat{X}$, $\mat{Z}$ and $\mat{Z}$, we arrive at    
\begin{equation}
[\nabla\tnr{\psi}(\tnr{B})] = 2([\tnr{B}]\mat{Z} - \mat{Y}\mat{X})\,.
\end{equation}
Therefore, the GD iterative equality \eqref{eq:gdm} arranged in matrix form for the case of function $\tnr{\psi}$ can be written as
\begin{equation}
[\tnr{B}_{t+1}] = [\tnr{B}_{t}] - 2\gamma_t([\tnr{B}_t]\mat{Z} - \mat{Y}\mat{X})\,,
\end{equation}
while the learning rate \eqref{eq:bbm} results
\begin{equation}\label{eq:learningRate}
\gamma_t = \dfrac{|\mat{L}:\mat{L}\mat{Z}|}{2\|\mat{L}\mat{Z}\|^2}\,,
\end{equation}
where $\mat{L}:=[\tnr{B}_{t}]-[\tnr{B}_{t-1}]$. Therefore, from the second iteration onward ($t\geqslant 1$), the iterative calculation
\begin{equation}
[\tnr{B}_{t+1}] = [\tnr{B}_{t}] - \dfrac{|\mat{L}:\mat{L}\mat{Z}|}{\|\mat{L}\mat{Z}\|^2}([\tnr{B}_t]\mat{Z} - \mat{Y}\mat{X})
\end{equation}
is performed until a specified small tolerance 
\begin{equation}
\epsilon > \|[\tnr{B}_{t+1}] - [\tnr{B}_{t}]\|\,.
\end{equation}
Since matrices $\mat{X}$ and $\mat{Z}$ are built from all the samples of the training set, this iterative method is also known as \emph{Stochastic Gradient Descent} (SGD). At this point, we have developed all the necessary information to propose another simple algorithmic solution for finding $\tnr{B}$, presented as Algorithm 2. 

\begin{algorithm}
\caption{Iterative calculation of tensor $\tnr{B}$ via GD}
\KwData{Training set $\pmb{\mathcal{T}}$}
\KwResult{$[\tnr{B}]$}
Specify $\epsilon$\;
Guess $\gamma_0,[\tnr{B}_{0}]$\;
$\mat{X} \gets \begin{bmatrix}
u^k_1 & \cdots & u^k_n & 1\\
\end{bmatrix}$ stacked vertically\;
$\mat{Y} \gets \begin{bmatrix}
{f}_1({\vto{u}}^k) & \cdots & {f}_m({\vto{u}}^k)
\end{bmatrix}^T$ stacked horizontally\;
$\mat{Z} \gets \mat{X}^T\mat{X}$\;
$\mat{K} \gets \mat{Y}\mat{X}$\;
$[\tnr{B}_{1}] \gets [\tnr{B}_{0}] - 2\gamma_0([\tnr{B}_0]\mat{Z} - \mat{K})$\;
$\mat{L}\gets[\tnr{B}_{1}]-[\tnr{B}_{0}]$\;
$t\gets 1$\;
\While{$\|\mat{L}\| > \epsilon$}{
$[\tnr{B}_{t+1}] \gets [\tnr{B}_{t}] - ([\tnr{B}_t]\mat{Z} - \mat{K})|\mat{L}:\mat{L}\mat{Z}|/\|\mat{L}\mat{Z}\|^2$\;
$\mat{L}\gets[\tnr{B}_{t+1}]-[\tnr{B}_{t}]$\;
$t\gets t+1$\;
}
\end{algorithm}

However, an important question still remains: how do we specify a good guess for the initial values $\tnr{B}_{0}$ and $\gamma_0$? A convenient strategy to avoid inconsistent guesses, which delays the convergence or even make it unattainable, is to specify values consistent with the dimensionality of the magnitudes involved. Let $\dim \mat{M}$ be the unit of measure (dimension) of a matrix $\mat{M}$, corresponding to the unit of measure of its elements. Note that this definition is adequate only if target and feature coordinates are normalized, that is, if features and targets groups are each one numerically comparable. In this context, from equations $\eqref{eq:analytic}$ and \eqref{eq:learningRate}, we can easily conclude that  
\begin{alignat}{3} 
	\dim [\tnr{B}] = \dim \mat{Y} (\dim \mat{X})^{-1}&\qquad \text{and}\qquad & \dim \gamma = (\dim \mat{X})^{-2}\,.
\end{alignat}
From these equalities, an example of a dimensional consistent guesses for $\tnr{B}_{0}$ and $\gamma_0$ could be
\begin{alignat}{3} 
	[\tnr{B}_0]_{ij} = \dfrac{\|\mat{Y}\|}{\|\mat{X}\|}&\qquad \text{and}\qquad &  \gamma_0 = \dfrac{1}{2\|\mat{X}\|^2}\,.
\end{alignat}

\section{Artificial Neural Network Regression}

We recall some of the conditions presented in the previous section, namely, the subspaces $\tnr{U}$ and $\tnr{V}$ of the respective real Hilbert spaces $\tnr{\mathcal{U}}^{n+1}$ and $\tnr{\mathcal{V}}^{m}$, where $\dim(\tnr{U})=n$ and $\dim(\tnr{V})=m$ belong to $\nat^*$. The regression problem we want to solve is also defined by the training set 
\begin{equation}\label{eq:train}
\pmb{\mathcal{T}}=\{ (\vto{u}^1,\vtf{f}(\vto{u}^1)),\cdots, (\vto{u}^p,\vtf{f}(\vto{u}^p))\}\,,\, p\in\nat^*,
\end{equation}
where $\map{\vtf{f}}{\tnr{U}}{\tnr{V}}$, but here the approximation function $\widetilde{\vtf{f}}$ can be non-linear. Considering $\tnr{Z}^{(1)}$ a $q_1$-dimensional subspace of real Hilbert space $\tnr{\mathcal{Z}}^{q_1}$, $ q_1\in\nat^*$, the \emph{Artificial Neural Network} (ANN)  model, in its simplest form, proposes to obtain this approximation function by defining an \emph{intermediary} or \emph{hidden} mapping $\map{\widetilde{\vtf{f}}^{(1)}}{\tnr{U}}{\tnr{Z}^{(1)}}$ and a \emph{final} mapping
$\map{\vtf{f}^{(2)}}{\tnr{Z}^{(1)}}{\tnr{V}}$ such that 
\begin{equation}\label{eq:composition}
\widetilde{\vtf{f}} = \widetilde{\vtf{f}}^{(2)}\circ\widetilde{\vtf{f}}^{(1)}\,.
\end{equation}
The method then defines that the rules of the hidden and final vector functions have the form
\begin{equation}\label{eq:layeri}
	\widetilde{\vtf{f}}^{(l)}(\vto{x})=\vtf{\phi}^{(l)}(\tnr{W}^{(l)}\odot{\vto{x}}+\vto{b}^{(l)})\,, l\in 1,2\,,
\end{equation}
where $\tnr{W}^{(1)}\in\tnr{Z}^{(1)}\otimes\tnr{U}$ and $\tnr{W}^{(2)}\in\tnr{V}\otimes\tnr{Z}^{(1)}$ are called the \emph{weight tensors}. The vector operators in $\map{\vtf{\phi}^{(1)}}{\tnr{Z}^{(1)}}{\tnr{Z}^{(1)}}$ and $\map{\vtf{\phi}^{(2)}}{\tnr{V}}{\tnr{V}}$, called \emph{activation functions}, must be differentiable and are always specified. The rule of the approximation function is
\begin{equation}\label{eq:approxi}
	\widetilde{\vtf{f}}(\vto{x})=\vtf{\phi}^{(2)}(\tnr{W}^{(2)}\odot[\vtf{\phi}^{(1)}(\tnr{W}^{(1)}\odot{\vto{x}}+\vto{b}^{(1)})]+\vto{b}^{(2)})\,,
\end{equation}
from which we conclude that the unknowns of the ANN model are the weight tensors and the \emph{bias} vectors $\vto{b}^{(1)}\in\tnr{Z}^{(1)}$ and $\vto{b}^{(2)}\in\tnr{V}$. The sets of coordinate function values
\begin{alignat}{3} 
	\tnr{L}_1:=\{ \widetilde{{f}}^{(1)}_1(\vto{x}),\cdots, \widetilde{{f}}^{(1)}_{q_1}(\vto{x})\} &\quad \text{and}\quad &  \tnr{L}_2:=\{ \widetilde{{f}}^{(2)}_1(\vto{x}),\cdots, \widetilde{{f}}^{(2)}_{m}(\vto{x})\} 
\end{alignat}
are called \emph{layers} and each of their elements are said to be \emph{neurons} or \emph{units}. The last layer $\pmb{{L}}_2$ is commonly called the \emph{target} or \emph{output layer} while $\pmb{{L}}_1$ is classified as hidden. Although the set $\{\vto{u}^1,\cdots,\vto{u}^p\}$, which is the input data for $\widetilde{\vtf{f}}^{(1)}$, is not formally defined as a layer, it is a common practice to call it an ``input layer''. The simplest ANN model, usually called \emph{shallow} ANN, has then two layers: one hidden and the output. Moreover, when none of the coordinates of the weight tensor $\tnr{W}^{(l)}$ is specified to be a constant, generally zero, layer $l$ is said to be \emph{fully connected} or \emph{dense} because in this case every unit of layer $l$ depends on every unit of $l-1$. 

For regression ANN models, by considering the activation functions  
\begin{equation}\label{eq:activation}
	\vtf{\phi}^{(l)}(\vto{x}) = \sum_{i=1}^{q_l} \phi^{(l)}(x_i)\vun{z}_i\,,
\end{equation}
where $q_2=m$ and $\{\vun{z}_1,\cdots,\vun{z}_{q_l}\}$ is an orthonormal basis of $\tnr{Z}^{(1)}$, scalar functions $\phi_i^{(l)}$ are usually specified among the most commonly used below. 
\begin{itemize}
	\item [i.] Sigmoid: given $\map{\sigma}{\real}{(0,1)}$\,,
	\begin{equation}
		\sigma(x)=\dfrac{1}{1+e^x}\,;
	\end{equation}
	\item [ii] Recified Linear Unit (ReLU): given $\map{(\bullet)^+}{\real}{[0,+\infty)}$\,,
	\begin{equation}
		(x)^+=\max(0,x)\,;
	\end{equation}
	\item [iii.] Leaky ReLU: given $\beta\in\real$ small and $\map{(\bullet)_L^+}{\real}{(-\infty,+\infty)}$\,,
	\begin{equation}
		(x)_L^+= \begin{cases} x, & x > 0 \\ 
			\beta x, & x \leqslant 0 \end{cases}\,;
	\end{equation}
	\item [iv.] Parametric ReLU: considering $\beta\in\real$ unknown and $\map{(\bullet)_P^+}{\real}{(-\infty,+\infty)}$\,,
	\begin{equation}
		(x)_P^+= \begin{cases} x, & x > 0 \\ 
			\beta x, & x \leqslant 0 \end{cases}\,;
	\end{equation}
	\item [v.] Exponential Linear Unit (ELU): given $\beta\in\real$ and $\map{(\bullet)_E^+}{\real}{(-\infty,+\infty)}$\,,
	\begin{equation}
		(x)_E^+= \begin{cases} x, & x > 0 \\ 
			\beta (e^x-1), & x \leqslant 0 \end{cases}\,;
	\end{equation}
	\item [vi.] Swish: given $\map{s}{\real}{(-\infty,+\infty)}$\,,
	\begin{equation}
		s(x)=x\sigma(x)\,;
	\end{equation}
	\item [vii.] Softmax: given a $n$-dimensional vector $\vto{x}$ and $\map{softmax}{\real}{(0,1)}$\,,
	\begin{equation}
		softmax(x_i)=\dfrac{e^{x_i}}{\sum_{i=1}^ne^{x_i}}\,
	\end{equation}
	from which 	$\sum_{i=1}^nsoftmax(x_i)=1$.
\end{itemize}
The composite function of any two of the previous activation functions have greater complexity and non-linearity than each one of them alone. In practice, from composition \eqref{eq:composition}, an ANN with three layers is ``more'' non-linear  than a model with two layers and tends to fit better a ``more'' non-linear complex training set. Therefore, in order to increase the level of complexity and non-linearity of the model, let's consider an ANN with $k> 2$ number of layers and generalize \eqref{eq:composition} the following way:
\begin{equation}
	\widetilde{\vtf{f}} = \widetilde{\vtf{f}}^{(k)}\circ\cdots\circ\widetilde{\vtf{f}}^{(0)}\,,
\end{equation}
where $\widetilde{\vtf{f}}^{(0)}$ is the identity function whose domain is the input layer. In this condition, the model is said to be a \emph{deep} ANN and its level of deepness increases with the number of layers. From definition \eqref{eq:layeri}, the approximation function rule in this case results
\begin{equation}
	\widetilde{\vtf{f}}(\vto{x})=\vtf{\phi}^{(k)}(\tnr{W}^{(k)}\odot[\underbrace{\widetilde{\vtf{f}}^{(k-1)}\circ\cdots\circ\widetilde{\vtf{f}}^{(0)}(\vto{x})}_{\vto{z}^{(k-1)}}]+\vto{b}^{(k)})\,.
\end{equation}
Given $0<l\leqslant k$, the approximation function rule of an arbitrary layer $l$ can be generically described in terms of the feature domain $\tnr{U}$ as 
\begin{equation}\label{eq:genericLayer}
	\widetilde{\vtf{f}}^{(l)}(\vto{x})=\vtf{\phi}^{(l)}(\tnr{W}^{(l)}\odot[\underbrace{\widetilde{\vtf{f}}^{(l-1)}\circ\cdots\circ\widetilde{\vtf{f}}^{(0)}(\vto{x})}_{\vto{z}^{(l-1)}}]+\vto{b}^{(l)})\,,
\end{equation}
where weight tensors $\tnr{W}^{(l)}\in\tnr{Z}^{(l)}\otimes\tnr{Z}^{(l-1)}$. Similarly to the previous section, the ANN model proposes criteria for accepting $\widetilde{\vtf{f}}$ based on a differentiable mapping  $\map{\tnr{\psi}^{(k)}}{\tnr{V}\otimes\tnr{Z}^{(k-1)}\times\tnr{V}}{\real}$ whose rule is 
\begin{equation}\label{eq:mean_error}
	\tnr{\psi}^{(k)}(\tnr{X},\vto{y})=\dfrac{1}{p}\sum_{i=1}^{p}\tnr{\mathcal{L}}[\vtf{\phi}^{(k)}(\tnr{X}\odot\vto{z}^{(k-1)}_i+\vto{y}),\vtf{f}(\vto{u}^i)]\,,
\end{equation}
where $\tnr{\psi}^{(k)}$ is called \emph{batch loss function}, argument $\vtf{\phi}^{(k)}(\tnr{X}\odot\vto{z}^{(k-1)}_i+\vto{y}):=\widetilde{\vtf{f}}(\vto{u}^i)$ and $\tnr{\mathcal{L}}$ is a scalar-valued vector function called \emph{loss function}. In this context, the typical loss functions rules used in ANN regression models are presented below.
\begin{itemize}
	\item [i.] Mean Squared Error: 
	\begin{equation}
		MSE(\vto{x},\vto{y})=\dfrac{1}{m}\|\vto{x}-\vto{y}\|^2\,;
	\end{equation}
	\item [ii.] Mean Absolute Error: 
	\begin{equation}
		MAE(\vto{x},\vto{y})=\dfrac{1}{m}\sum_{i=1}^m|x_i-y_i|\,;
	\end{equation}
	\item [iii.] Huber: given a positive scalar parameter $\delta$,
	\begin{equation}
		H_\delta(\vto{x},\vto{y})=\sum_{i=1}^m\begin{cases} (x_i-y_i)^2/2, & |x_i-y_i| \leqslant \delta \\ 
			\delta|x_i-y_i| - \delta^2/2, & |x_i-y_i| > \delta \end{cases}\,;
	\end{equation}
	\item [iv.] Log-Cosh: 
	\begin{equation}
		LCH(\vto{x},\vto{y})=\dfrac{1}{m}\sum_{i=1}^m\log(\cosh(x_i-y_i)\,;
	\end{equation}
	\item [v.] Mean Squared Logarithmic Error: 
	\begin{equation}
		MSLE(\vto{x},\vto{y})=\dfrac{1}{m}\sum_{i=1}^m[\log(x_i+1)-\log(y_i+1)]^2\,;
	\end{equation}
	\item [vi.] Poisson: 
	\begin{equation}
		POI(\vto{x},\vto{y})=\dfrac{1}{m}\sum_{i=1}^m x_i-x_i\log(y_i)\,.
	\end{equation}
\end{itemize}

\paragraph{The Backpropagation Algorithm.} In this subsection, we describe an iterative strategy to obtain a suitable approximation function $\widetilde{\vtf{f}}$ by first guessing weights and biases to forwardly calculate unit values from the first hidden layer to the output layer and then backwardly propagate from the output to the first hidden layer the gradient of the batch loss $\tnr{\psi}^{(k)}$ on the output layer. The goal of this backward propagation is to compute the contribution of each layer to the value of the batch loss.  

For an arbitrary pair $(\vto{u}^i,\vtf{f}(\vto{u}^i))$ of the training set, the first forward pass of the Backpropagation algorithm is performed by guessing weights and biases from layer $1$ to layer $k$ in order to obtain all the unit values according to 
\begin{equation}
	\widetilde{\vtf{f}}(\vto{u}^i)=\vtf{\phi}^{(k)}(\tnr{W}^{(k)}\odot[\widetilde{\vtf{f}}^{(k-1)}\circ\cdots\circ\widetilde{\vtf{f}}^{(0)}(\vto{u}^i)]+\vto{b}^{(k)})\,.
\end{equation}
Still considering the arbitrary pair above and a variable 
\begin{equation}\label{eq:variab}
	\vto{z} = \tnr{X}\odot\vto{z}^{(k-1)}_i+\vto{y}\,,
\end{equation}
from the right hand side of rule \eqref{eq:mean_error}, we can write that 
\begin{equation}\label{eq:eqLoss}
\tnr{\mathcal{L}}[\vtf{\phi}^{(k)}(\vto{z}),\vtf{f}(\vto{u}^i)]=\tnr{\mathcal{\hat{L}}}_i\circ\vtf{\phi}^{(k)}(\vto{z})\,,
\end{equation}
where $\tnr{\mathcal{\hat{L}}}_i(\vto{x})=\tnr{\mathcal{L}}[\vto{x},\vtf{f}(\vto{u}^i)]$. Tensor Calculus teaches us that the gradient of the loss function above follows the chain rule for gradients and then 
\begin{equation}
\nabla(\tnr{\mathcal{\hat{L}}}_i\circ\vtf{\phi}^{(k)})(\vto{z})=\nabla\vtf{\phi}^{(k)}(\vto{z})\odot\nabla\tnr{\mathcal{\hat{L}}}_i\circ\vtf{\phi}^{(k)}(\vto{z})\,.
\end{equation}
After defining the loss function $\tnr{\mathcal{L}}$, since $\vto{z}_i^{(k)}:=\widetilde{\vtf{f}}^{(k)}(\vto{u}^i)$ is known from the forward pass, it is possible to calculate the vector
\begin{equation}\label{eq:deltaK}
\tnr{\delta}^{(k)}_i:=\nabla(\tnr{\mathcal{\hat{L}}}_i\circ\vtf{\phi}^{(k)})(\vto{z}_i^{(k)})\,.
\end{equation}
Considering the same equality as above for the hidden layers $r=k-1,\cdots,1$, the algorithm specifies that
\begin{equation}
  \nabla\tnr{\mathcal{\hat{L}}}_i\circ\vtf{\phi}^{(r)}(\vto{z}_i^{(r)})=\tnr{\delta}^{(r+1)}_i\odot{\tnr{W}^{(r+1)}}\,,
\end{equation}
where the weight tensor $\tnr{W}^{(r+1)}$ is known from the forward pass. This calculation for $r=k-1$ marks the beginning of the backward pass of the algorithm, from which its name derives. Considering the previous definition,
\begin{equation}\label{deltaR}
  \tnr{\delta}^{(r)}_i=\nabla\vtf{\phi}^{(r)}(\vto{z}_i^{(r)})\odot (\tnr{\delta}^{(r+1)}_i\odot{\tnr{W}^{(r+1)}})\,.
\end{equation}
From the variable in \eqref{eq:variab} and equality \eqref{eq:eqLoss}, it is possible to rewrite \eqref{eq:mean_error} the following way:
\begin{equation}
\tnr{\psi}^{(r)}(\tnr{X},\vto{y})=\dfrac{1}{p}\sum_{i=1}^{p}\tnr{\mathcal{\hat{L}}}_i\circ\vtf{\phi}^{(r)}(\tnr{X}\odot\vto{z}^{(r-1)}_i+\vto{y})\,,
\end{equation}
from which, by applying the chain rule and the product rule for gradients, it is possible to arrive at constant functions
\begin{equation}\label{eq:gradX}
\nabla_{\tnr{X}}\tnr{\psi}^{(r)}(\tnr{X},\vto{y})=\nabla_{\tnr{X}}\tnr{\psi}^{(r)}=\dfrac{1}{p}\sum_{i=1}^{p} \tnr{\delta}^{(r)}_i\odot\tnr{I}^4\odot\vto{z}^{(r-1)}_i\,,
\end{equation}
where $\tnr{I}^4\in\tnr{Z}^{(r)}\otimes\tnr{Z}^{(r-1)}\otimes\tnr{Z}^{(r)}\otimes\tnr{Z}^{(r-1)}$ is an identity tensor described by \eqref{eq:identity}, and 
\begin{equation}\label{eq:gradY}
\nabla_{\vto{y}}\tnr{\psi}^{(r)}(\tnr{X},\vto{y})=\nabla_{\vto{y}}\tnr{\psi}^{(r)}=\dfrac{1}{p}\sum_{i=1}^{p}\tnr{\delta}^{(r)}_i\textbf{}\,.
\end{equation}
At this point, we adjust weights and biases of the layer $r$ based on their previous values and the calculated gradients above by specifying that
\begin{alignat}{3} 
\tnr{W}^{(r)}_{t+1}=\tnr{\omega}_{\tnr{W}}(\nabla_{\tnr{X}}\tnr{\psi}^{(r)},\tnr{W}^{(r)}_t) &\quad \text{and}\quad &  \vto{b}^{(r)}_{t+1}=\tnr{\omega}_{\vto{b}}(\nabla_{\vto{y}}\tnr{\psi}^{(r)},\vto{b}^{(r)}_t)\,,
\end{alignat}
where functions $\tnr{\omega}_{\tnr{W}}$ and $\tnr{\omega}_{\vto{b}}$ are called \emph{optimizers}. Given a learning rate $\gamma$, the gradient descent is an example of optimizer:   
\begin{alignat}{3}\label{eq:annGd} 
		\tnr{W}^{(r)}_{t+1}=\tnr{W}^{(r)}_t-\gamma\nabla_{\tnr{X}}\tnr{\psi}^{(r)}
	 &\quad \text{and}\quad &  \vto{b}^{(r)}_{t+1}=\vto{b}^{(r)}_{t}-\gamma\nabla_{\vto{y}}\tnr{\psi}^{(r)}\,.
\end{alignat}
Among the various optimizers that were created based on the gradient descent, the most important are the following:
\begin{itemize}
	\item [i.] Momentum: accelerates the gradient descent by adding a fraction of the update vector of the past time step to the current update.
	\item [ii.] Nesterov: variant of Momentum that anticipates the future position of the parameters;
	\item [iii.] Adaptative Gradient: adapts the learning rate of the GD for each parameter individually;
	\item [iv.] Root Mean Square Propagation (RMSProp): adapts the Adaptative Gradient to prevent learning rate from decreasing too much;
	\item [v.] Adaptative Moment Estimation (Adam): combines the advantages of RMSProp and Momentum;
	\item [vi.] Nesterov Adaptative Moment Estimation (Nadam): combines the advantages of Nesterov and Adam.
\end{itemize}
After the weights and biases in \eqref{eq:annGd} are calculated, a new forward process begins, that is, a new iteration of the algorithm is performed. Every time the entire training set is processed is called an \emph{epoch}. In a \emph{batch mode} of processing, which is our case here, where the entire training set is submitted to every new iteration of the algorithm, the number of iterations and epochs are the same. When batch mode is not feasible due to the large size of the training set, we commonly use a method of dividing this set into smaller sets that are successively submitted to every new iteration of the algorithm until an epoch is completed. In this mode of processing, called \emph{mini-batch}, the number of iterations and epochs can be different. 

Now, similarly to the training set, let us consider a validation set
\begin{equation}\label{eq:vali}
	\pmb{\mathcal{V}}=\{ (\vto{v}^1,\vtf{f}(\vto{v}^1)),\cdots, (\vto{v}^s,\vtf{f}(\vto{v}^s))\}\,,\, s\in\nat^*,
\end{equation}
and a loss value 
\begin{equation}
	\vartheta_e:=\dfrac{1}{s}\sum_{i=1}^{s}\tnr{\mathcal{L}}[\widetilde{\vtf{f}}^{(k)}(\vto{v}^i),\vtf{f}(\vto{v}^i)]
\end{equation}
that is calculated for each epoch $e$ at the end of a forward pass. Given a small positive tolerance $\epsilon$, the most common stopping criterion for the Backpropagation Algorithm is when one of the following conditions is obeyed first: a specified \emph{Maximum Number of Epochs} is reached or 
\begin{equation}\label{eq:tolerance}
	\epsilon > | \vartheta_e - \vartheta_{e-1} |\,.
\end{equation}   

Still considering the batch mode of processing and $0<l\leqslant k$, for an arbitrary layer $l$, we specify weight and bias matrices for all the samples of the training and validation sets as the following:   
\begin{equation}
	\mat{W}^{(l)}:=\begin{bmatrix}
		{W}^{(l)}_{11} & \cdots & {W}^{(l)}_{1q_{l-1}} \\
		\vdots & \ddots & \vdots \\
		{W}^{(l)}_{q_l1} & \cdots & {W}^{(l)}_{q_lq_{l-1}} 
	\end{bmatrix}
\end{equation}
and 
\begin{equation}
\mat{B}^{(l)}:=\underbrace{\begin{bmatrix}
	b^{(l)}_1  & b^{(l)}_1  &\cdots & b^{(l)}_1 \\
	\vdots & \vdots &\ddots & \vdots  \\
	b^{(l)}_{q_l} & b^{(l)}_{q_l} &\cdots & b^{(l)}_{q_l} 
\end{bmatrix}}_{\text{values repeated }p+s\text{ times columnwise}}
\end{equation}
where $\mat{W}^{(l)}=[\tnr{W}^{(l)}]^T$ . Concerning the activation functions described by \eqref{eq:activation}, in order to simplify notation, for a given matrix $\mat{A}$, we define a matrix $\phi^{(l)}(\mat{A})$ such that 
\begin{equation}
	(\phi^{(l)}(\mat{A}))_{ij} = \phi^{(l)}(\mat{A}_{ij})\,.
\end{equation}
Generic equation \eqref{eq:genericLayer} can then be rewritten in matrix terms as
	\begin{equation}\label{eq:matUnit}
	\mat{Z}^{(l)}= \phi^{(l)}(\mat{W}^{(l)}\mat{Z}^{(l-1)} + \mat{B}^{(l)})\,,
\end{equation}
where the matrix 
\begin{equation}
	\mat{Z}^{(0)}:= \begin{bmatrix}
		{u}^1_1 & \cdots & {u}^p_1 & {v}^1_1 & \cdots & {v}^s_1\\
		\vdots & \ddots & \vdots & \vdots & \ddots & \vdots\\
		{u}^1_n & \cdots & {u}^p_n & {v}^1_n & \cdots & {v}^s_n
	\end{bmatrix}\,.
\end{equation}

Once $\mat{Z}^{(k)}$ is calculated and the loss function defined, from definition \eqref{eq:deltaK}, it is possible to specify 
\begin{equation}
	\Delta^{(k)}:= \begin{bmatrix}
		{(\tnr{\delta}^{(k)}_1)}_1 & \cdots & {(\tnr{\delta}^{(k)}_p)}_1 \\
		\vdots & \ddots & \vdots \\
		{(\tnr{\delta}^{(k)}_1)}_m & \cdots & {(\tnr{\delta}^{(k)}_p)}_m 
	\end{bmatrix}\,,
\end{equation}
from which and equality \eqref{deltaR}, matrices $\Delta^{(r)}$ can be calculated backwardly for each layer $r$. From these matrices and equalities \eqref{eq:gradX} and \eqref{eq:gradY}, matrices $[\nabla_{\tnr{X}}\tnr{\psi}^{(r)}]$  and  $[\nabla_{\vto{y}}\tnr{\psi}^{(r)}]$ can be obtained. Then, by choosing an optimizer, we get matrices $\mat{W}^{(r)}_{t+1}$ and $\mat{B}^{(r)}_{t+1}$ until $r=1$. At the end of each epoch, it is possible to calculate $\vartheta_e$ from $\mat{Z}^{(k)}$ and inspect \eqref{eq:tolerance}. Algorithm 3 presents this procedure step by step.

Just like the OLS model, in ANN, initial guesses of the unknowns are also a crucial step for the convergence of the iterative process, not only in terms of speed but in also in terms of the quality of the solution. Biases are typically initialized with zero or with small positive values near zero. For the case of weights, there are the following approaches:  
\begin{itemize}
	\item [i.] Random Guess: random numbers are generated from an uniform distribution within a range $(-\beta,+\beta)$ or from a normal distribution within $(0,\sigma^2)$.
	\item [ii.] Xavier-Glorot Guess: random numbers are generated from an uniform distribution within a range $(-\sqrt{6}/(\sqrt{q_{i-1}} + \sqrt{q_i}),\sqrt{6}/(\sqrt{q_{i-1}} + \sqrt{q_i})$.
	\item [iii.] Kaiming Guess: random numbers are generated from an uniform distribution within a range $(-\sqrt{6/q_{i-1}},\sqrt{6/q_{i-1}})$ and normal distribution within $(0,\sqrt{2/q_{i-1}})$.
	\item [iv.] LeCun Guess: random numbers are generated from a normal distribution within $(0,\,1/q_{i-1})$.
\end{itemize}

\begin{algorithm}
	\SetNoFillComment
	\caption{Backpropagation Algorithm in Batch Mode}
	\KwData{Training set $\pmb{\mathcal{T}}$}
	\KwData{Validation set $\pmb{\mathcal{V}}$}
	\KwResult{$\{(\mat{W}^{(1)},\mat{B}^{(1)}),\cdots,(\mat{W}^{(k)},\mat{B}^{(k)})\}$}
	Specify $k$, $\{q_1,\cdots,q_{k-1}\}$, $\{{\phi}^{(1)},\cdots,{\phi}^{(k)}\}$, $\tnr{\mathcal{L}}$, $\tnr{\omega}$\;
	Specify maximum number $\alpha$ of epochs and tolerance $\epsilon$\; 
	Guess  $\{(\mat{W}^{(1)}_1,\mat{B}^{(1)}_1),\cdots,(\mat{W}^{(k)}_1,\mat{B}^{(k)}_1)\}$\;
	Build $\mat{Z}^{(0)}$\; 
	$c\gets MAX\_FLOAT$\;
	$\vartheta_0\gets MAX\_FLOAT$\;
	$t\gets 1$\;
	\While{$c > \epsilon$ and $t\leqslant \alpha$}{
		$l\gets 1$\;
		\tcc{Forward pass}
		\While{$l\leqslant k$}{
		$\mat{Z}^{(l)}_t\gets \phi^{(l)}(\mat{W}^{(l)}_t\mat{Z}^{(l-1)}_t + \mat{B}^{(l)}_t)$\;
		$l\gets l+1$\;
		}
	    Calculate $\vartheta_t $ from $\mat{Z}^{(k)}_t$\;
	    $c\gets |\vartheta_t-\vartheta_{t-1}|$\;
	    \If{$c \geqslant \epsilon$}{
	    		Calculate $\Delta^{(k)}_t$ from $\mat{Z}^{(k)}_t$ and $\tnr{\mathcal{L}}$\;
	    		$r\gets k-1$\;
	    		\tcc{Backward pass}
	    		\While{$r > 0$}{
	    			Calculate $\Delta^{(r)}_t$ from $\Delta^{(r+1)}_t$ and $\mat{W}^{(r+1)}_t$\;
	    			Calculate $[\nabla_{\tnr{X}}\tnr{\psi}^{(r)}]_t$ and $[\nabla_{\vto{y}}\tnr{\psi}^{(r)}]_t$ from $\Delta^{(r)}_t$\;
	    			Calculate $\mat{W}^{(r)}_{t+1}$ and $\mat{B}^{(r)}_{t+1}$ from $\tnr{\omega}$\;
	    			$r\gets r -1$\;
	    		}
	    }
		$t\gets t+1$\;
	}
\end{algorithm}

\end{document}